\documentclass[a4paper,11pt,onecolumn]{scrartcl}
\usepackage{graphicx}
\usepackage{amsmath,amssymb} 
\usepackage{color}
\usepackage{nicefrac}
\usepackage[pdftex,
            pdfauthor={Dominik A. Klein, Dirk Schulz, Armin B. Cremers},
            pdftitle={Realtime Hierarchical Clustering based on Boundary and Surface Statistics},
            pdfsubject={Computer Vision for Robotics},
            pdfkeywords={},
            pdfproducer={},
            pdfcreator={},
            pagebackref=true,breaklinks=true,letterpaper=false,colorlinks,bookmarks=false]{hyperref}
\usepackage{fancyhdr}

\newcommand{\radigabs}[1]{
  \ensuremath{\left\vert #1 \right\vert}
}

\newcommand{\radigtvec}[3]{
  \ensuremath{\left(\begin{smallmatrix}\text{#1}\\\text{#2}\\\text{#3}\end{smallmatrix}\right)}%
}

\newcommand{\radigwdist}{\ensuremath{\operatorname{\mathcal{W}}_1}}
\newcommand{\radigtrace}[1]{\ensuremath{\operatorname{tr}\left(#1\right)}}

\begin{document}
\pagestyle{fancy}
\lhead{ACCV 2016}
\rhead{The final publication is available at \url{link.springer.com}}



\title{Realtime Hierarchical Clustering based on Boundary and Surface Statistics} 

\author{Dominik Alexander Klein$^1$, Dirk Schulz$^1$, and Armin Bernd Cremers$^2$\\
\begin{small}
 $^1$ Fraunhofer FKIE, Dept. Cognitive Mobile Systems
\end{small}\\
\begin{small}
 $^2$ Bonn-Aachen Int. Center for Information Technology (b-it)
\end{small}}
\date{May 27, 2016}

\maketitle

\begin{abstract}
  Visual grouping is a key mechanism in human scene perception. There, it
  belongs to the subconscious, early processing and is key prerequisite for other
  high level tasks such as recognition. In this paper, we introduce an efficient,
  realtime capable algorithm which likewise agglomerates a valuable hierarchical
  clustering of a scene, while using purely local appearance statistics.
  
  To speed up the processing, first we subdivide the image into meaningful, atomic
  segments using a fast Watershed transform.
  Starting from there, our rapid, agglomerative clustering algorithm prunes and
  maintains the connectivity graph between clusters to contain only such pairs,
  which directly touch in the image domain and are reciprocal nearest neighbors
  (RNN) wrt. a distance metric.
  The core of this approach is our novel cluster distance: it combines boundary and
  surface statistics both in terms of appearance as well as spatial linkage.
  This yields state-of-the-art performance, as we demonstrate in conclusive experiments
  conducted on BSDS500 and Pascal-Context datasets.
\end{abstract}

\section{Introduction}
\label{sec:intro}
  One of the major challenges in computer vision is the question of semantic
  image partitioning. While today's cameras do record impressive numbers of
  pixels per image, the amount of possible segmentations and subdivisions of
  such pictures is even incredibly much higher. Therefore, generation of
  coherent parts and object candidates is a crucial step in every vision
  processing pipeline prior to higher level semantic interpretation.
  
  There are several algorithmic variants how to break down the data for
  semantic classification and object detection:
  For instance, the family of plain window scanning methods has become popular
  in classification of certain object types~\cite{violajones04,dalaltriggs05,Felzenszwalb10}.
  Usually, a vast amount of candidates is sampled regularly without respect
  to the image content itself. Still, such methods have issues generating the
  proper candidates, e.g.\ with objects that do not fit well in rectangular shapes.  
\begin{figure}[t]
\begin{center}
  \includegraphics[width=0.19\linewidth]{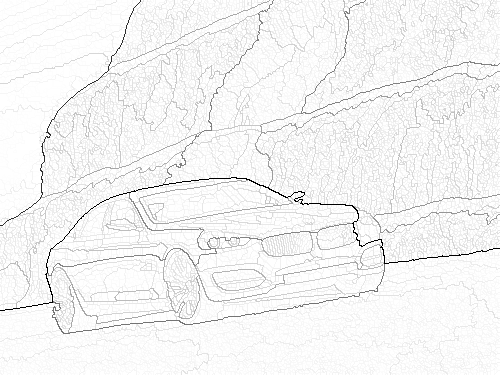}
  \includegraphics[width=0.19\linewidth]{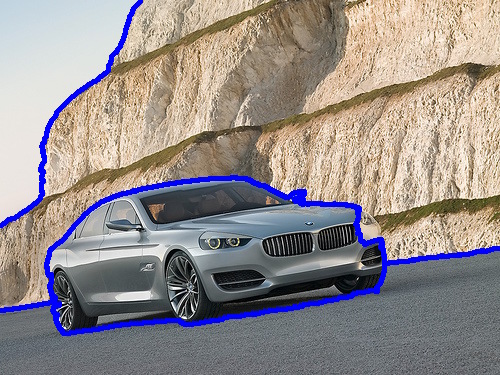}
  \includegraphics[width=0.19\linewidth]{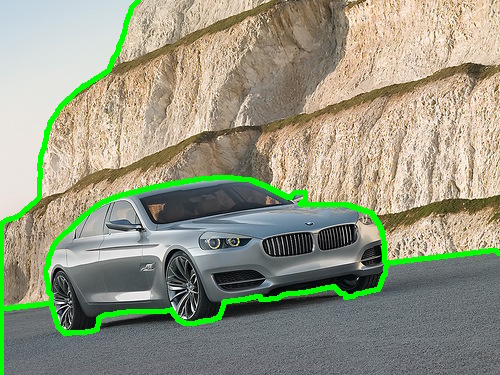}
  \includegraphics[width=0.19\linewidth]{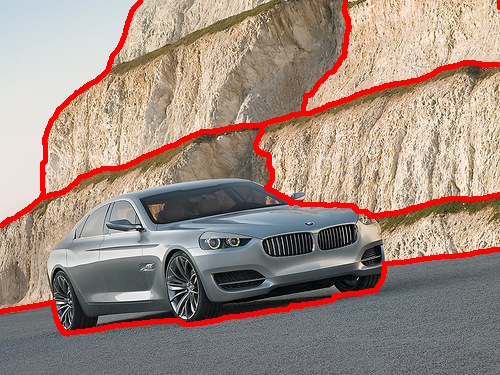}
  \includegraphics[width=0.19\linewidth]{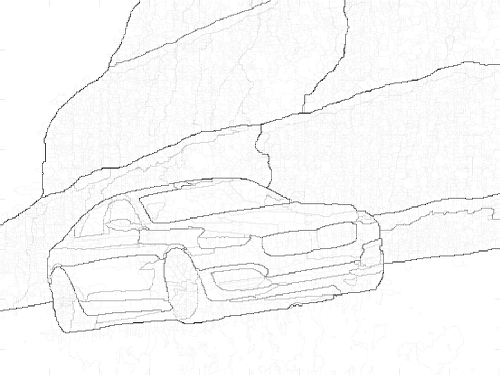}\\
  \includegraphics[width=0.19\linewidth]{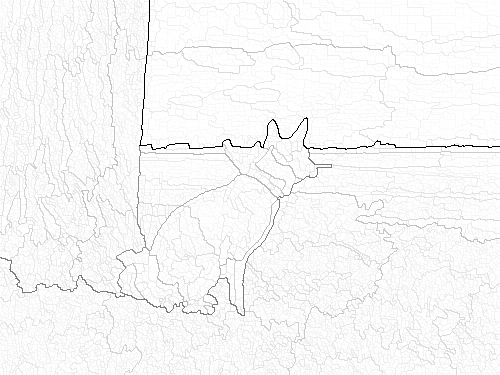}
  \includegraphics[width=0.19\linewidth]{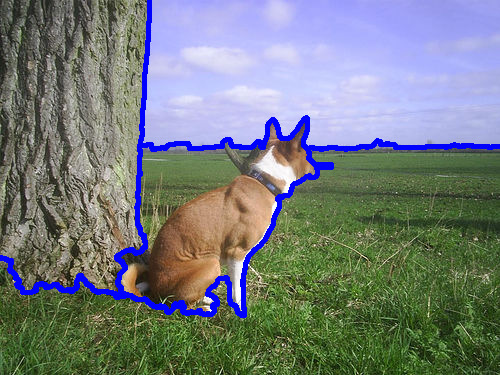}
  \includegraphics[width=0.19\linewidth]{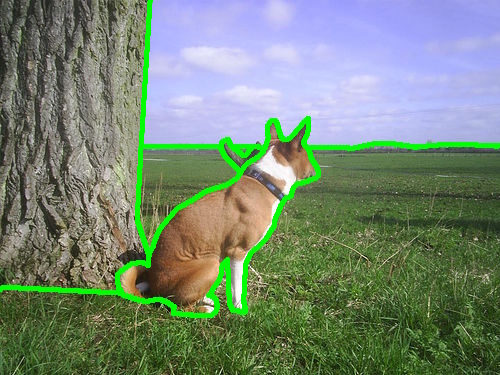}
  \includegraphics[width=0.19\linewidth]{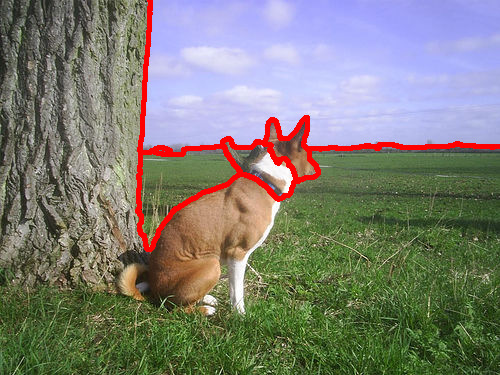}
  \includegraphics[width=0.19\linewidth]{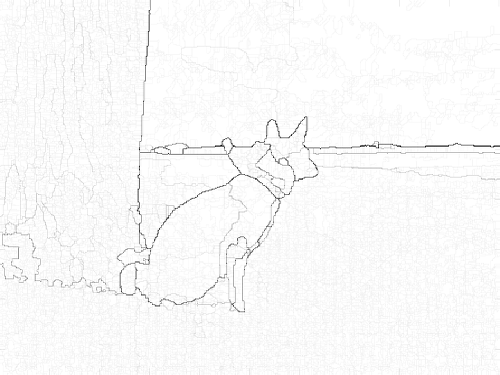}\vspace{-4mm}
\end{center}
   \caption{Exemplary results from Pascal-Context dataset~\cite{mottaghi2014role}. From
   left to right per row: ultrametric contour map (UCM) and optimal image scale (OIS)
   segmentation of our approach RaDiG (blue), ground-truth boundaries (green), OIS and UCM of the leading approach MCG-UCM~\cite{arbelaez2014mcg}(red).}
\label{fig:teaser}
\end{figure}
  A second group of candidate generating methods is based on interest
  point operators, which first locate prominent points before aligning
  known shapes with respect to matched ones~\cite{lowe1999object,sivic2003video,leibe2004combined}.
  This usually generates less and high quality object candidates. However, it
  requires an object model to transform the set of localized interest points
  into object candidates.
  This is a significant drawback when dealing with a large number of
  different part- and object classes or priorly unknown ones to be found.
  Unsupervised clustering is a third line in this taxonomy of approaches
  and the most biologically inspired one, since it resembles the mechanism of
  visual grouping, which has been explored by psychologists and neurophysiologists
  for decades~\cite{Goldstein2009,Wagemans2012}. Such methods can provide a complete partitioning of the scene
  arising from the data itself. Even more, some approaches yield a hierarchy of
  nested segmentations, which naturally corresponds to object-part relations.
  In essence, these methods rely on a proper distance metric to measure differences
  between image parts. The general hypothesis is that a semantically meaningful entity
  is of consistent appearance and in contrast to its surroundings in some way.
  A comprehensive research about what defines this \textit{consistency} in human
  perception was performed by the Gestalt school of psychology~\cite{Wertheimer12} and influenced
  many technical approaches.
  
  The algorithm introduced in this paper belongs to the unsupervised clustering
  approaches and is named RaDiG (\textbf{Ra}pid \textbf{Di}stributions \textbf{G}rouping). At its core,
  it follows the well known agglomerative clustering paradigm, greedily merging
  the pair of \textit{closest} segments in each step up to total unification, but innovating in many
  details. Our main contributions comprise
  \begin{itemize}
   \item an innovative, threefold distance metric incorporating boundary contrast with
   surface dissimilarity as well as spatial linkage,
   \item the texture-aware enhancement of Ward's minimum variance criterion
   with the meaningful and efficient Wasserstein distance in the space of normal distributions of features,
   \item a considerable reduction of runtime due to several algorithmic tweaks with
   pruning and maintaining the cluster connectivity graph as well as an economic
   feature processing.
  \end{itemize}
  In total, these improvements result in an expeditious algorithm which meets strict
  realtime requirements. Still, the quality of results is on level with the best,
  globally optimized and highly complex approaches (cf. Fig.~\ref{fig:teaser}). Therefore, we claim that it
  is the method of choice in mobile use cases such as robotics. This is
  supported by our experimental findings presented in sections~\ref{sec:eval_contrib}
  and~\ref{sec:eval_comp}.

\section{Related Work}
\label{sec:related}
  There is a huge history of research concerned with segmentation in general
  settings as well as more specific in computer vision. Here, we restrict our review to the most related, seminal
  and/or up-to-date methods, and in addition referring the reader to appropriate
  survey literature.
  Unsupervised clustering is a field at least as old as computer science itself.
  We recommend the textbook of Hastie, Tibshirani and Friedman~\cite{Hastie09} as
  a comprehensive work about algorithms and statistical background.
  More specialized on agglomerative hierarchical clustering and thus related to
  this paper is the overview written by Murtagh and Contreras~\cite{Murtagh12}.
  
  Image segmentation in computer vision is a wide subject with many applications.
  In this paper, we narrow our interest down to methods which segment natural
  images into semantically meaningful entities in a fully unsupervised fashion,
  i.e. without any prior knowledge about the image content or objects.\\  
  There are "planar" methods, which aim to find an optimal partitioning of the input
  image. An example is the famous mean shift approach~\cite{Comaniciu2002}, in which
  pixels are assigned to the next stationary point of an underlying density in the
  joint domain of color and image position.
  Some planar methods do integrate a notion of a best segmentation across all scales
  of structures in the image~\cite{Cour05,FelzHutt04}. Recently, the Superpixel
  paradigm became popular, which focuses on segmenting the basic building blocks of
  images~\cite{achanta2012slic,van2012seeds}. This usually attains an
  oversegmentation of semantic entities, therefore some approaches add a further
  accumulation step of Superpixels into larger clusters~\cite{kovesi13,zhouimage}.\\  
  Besides planar clustering techniques, there are methods providing a
  hierarchical tree (dendrogram) of nested segmentations. This additionally determined
  subset-relation can be useful for further processing.
  With such hierarchy, the scale  of results is not fixed: the nodes close to the root
  represent a coarse segmentation, which becomes a fine oversegmentation traversing
  the tree to the leaves~\cite{Peng13}.
  Most approaches can be categorized into either divisive or agglomerative type.
  Divisive ones build the hierarchy in a top-down manner: iteratively, the current
  partitioning is split into a finer one. For each step, one could apply a planar
  segmentation algorithm on each remaining cluster, until they show a uniform
  appearance. An advantage of divisive methods is that they can naturally exploit
  the statistics of an entire image/region to find an optimal splitting. Among such
  methods, there is the family of normalized cuts based
  approaches~\cite{Cour05,ShiMalik00}.
  In contrast, agglomerative approaches merge initial clusters in a bottom-up way
  until complete unification. Region based approaches maintain and match feature
  statistics of a segment's surface, such as the mean color or a histogram of
  texture~\cite{vilaplana2008,calderero2010,alpert2012image}.
  As Arbel\'aez~\cite{Arbelaez06} pointed out, a hierarchy tree is the dual
  representation of an ultrametric contour map (UCM), if the underlying metric holds
  the \textit{strong triangle inequality}. With this insight, the way of thinking
  image segmentation shifted to concentrate on contour detection. This fruitful
  ansatz led to a family of algorithms which globalize the results of sophisticated
  contour detectors using spectral methods, before a greedy agglomerative clustering
  constructs the hierarchy based on average boundary
  strengths~\cite{arbelaez2011contour,arbelaez2014mcg,Taylor13}.\\  
  Since a planar segmentation is achievable at lower computational complexity than
  hierarchical approaches, for the sake of processing speed it can be advantageous to
  first compute an initial oversegmentation, before constructing a hierarchy on
  top~\cite{haris1998,Marcotegui2005,Jain11}. In a similar way, contour based
  approaches benefit from an atomic oversegmentation since it can significantly
  accelerate the globalization of boundary strengths~\cite{Taylor13}. However, the
  spectral contour globalization process can be speed up alike by iterated decimation
  and information propagation of the affinity matrix~\cite{arbelaez2014mcg}.
  
  Our algorithm RaDiG also provides a hierarchy of segments for further
  processing and is designed to be realtime capable. Thus, we have chosen the greedy
  framework of agglomerative clustering starting from a topological Watershed
  transform. While this basis is similar to several other approaches, ours is
  considerably different in its components, such as the performant processing
  of the image graph structure, the efficient representation of feature distributions,
  and the profitable combination of novel boundary contrast, region dissimilarity
  as well as spatial linkage within an innovative cluster-distance.

\section{Rapid Distributions Grouping}
\label{sec:radig}
  \begin{figure}[t]
  \begin{center}
    \includegraphics[width=0.16\linewidth]{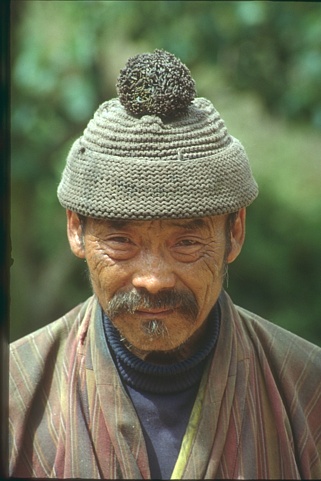}
    \includegraphics[width=0.16\linewidth]{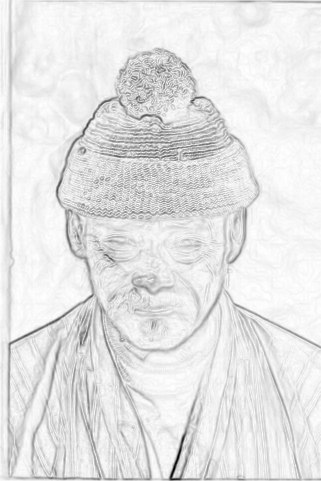}
    \includegraphics[width=0.16\linewidth]{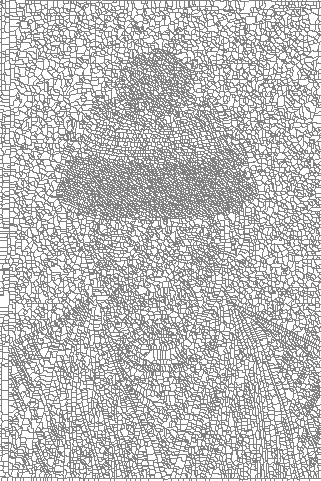}
    \includegraphics[width=0.16\linewidth]{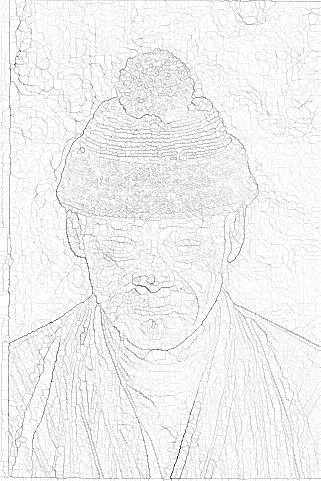}
    \includegraphics[width=0.16\linewidth]{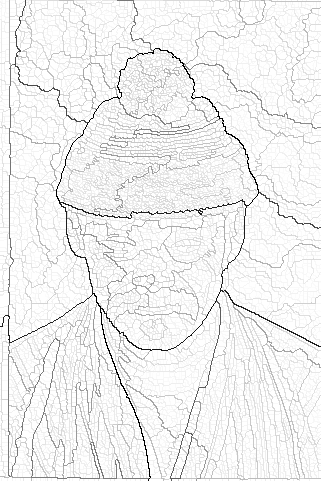}
    \includegraphics[width=0.16\linewidth]{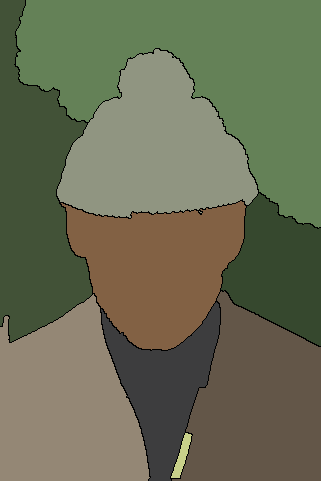}\vspace{-4mm}
  \end{center}
     \caption{RaDiG processing steps. From left to right: input image, gradient magnitude, watershed segments, initial cluster distances, final ultrametric contour map,
     optimal image scale (OIS) segmentation.}
  \label{fig:boundary_contrast}
  \end{figure}
  The execution of our method comprises four main steps. At first, we convert the
  colorspace of the input image to CIE-Lab. In the following sections, we will name the
  luminance dimension $L$ and the chromatic dimensions with $a$ resp. $b$. Next, we
  compute an oversegmentation of the image (Sec.~\ref{ssec:watershed}). This is used to
  initiate the finest layer of the hierarchical clustering (Sec.~\ref{ssec:init_tree}).
  Finally, our agglomerative approach greedily merges the pair of \textit{closest}
  segments in each step up to total unification (Sec.~\ref{ssec:agg_clustering}). Intermediate results of the approach are shown in Figure~\ref{fig:boundary_contrast}. All gained subset relations are kept in a tree structure to be available for further steps of a processing pipeline.

\subsection{Atomic Subdivision using Watershed Transform}
\label{ssec:watershed}
  Since the number of clusters is doubling with each additional level of the cluster
  hierarchy, it can potentially save a lot of processing time to prune the finest
  layers from the pixel level and replace them with a planar, atomic subdivision. An
  established method producing such oversegmentation based on irregularities in the
  image function is the watershed transform. We apply a variant of the topographical
  watershed transform based on hill climbing~\cite{roerdink2001} which features
  $\mathcal{O}(n)$ runtime complexity in the number of pixels and is very fast in
  practice. It operates on the gradient magnitude of the input image, which we
  determine as follows: the partial image derivatives in $x$ and $y$ directions
  are calculated by convolution with the optimized, derivative 5-tap filters of Farid
  and Simoncelli~\cite{farid2004} for each layer $L$, $a$, and $b$. Then, our $Lab$
  gradient magnitude is made of independent luminance and chromaticity parts
  \begin{equation}
   |\nabla| = \sqrt{L_x^2 + L_y^2} + \sqrt{2 (a_x^2 + a_y^2 + b_x^2 + b_y^2)} .
  \end{equation}
  There is a complete field of research on edge strength calculations by itself. While
  recently machine learning based methods have shown to improve bounding edge detections\cite{Xie2015,Yin2016}, that
  sophisticated and costly quantification is not crucial or often not even beneficial
  when initializing an oversegmentation.

\subsection{Founding of the Hierarchy Tree Structure}
\label{ssec:init_tree}
  \begin{table}[t]
   \def\arraystretch{1.5}
   \begin{scriptsize}
   \begin{center}
    \begin{tabular}{|l|l|}
     \hline
     \textbf{cluster} & \textit{hierar. connectivity}: parent, left-, and right child cluster-IDs \\
     \textbf{(node)} & \textit{planar connectivity}: adj.-list of \radigtvec{neighbor-ID}{boundary-ID}{cl.-distance}; near.-neighbor index\\
      & \textit{statistics}: area $A$ in number of pixels; 
      color distribution
      $\left\{\left(\begin{smallmatrix}\mu_L\\\mu_a\\\mu_b\end{smallmatrix}\right),
      \left(\begin{smallmatrix}\sigma_L\\\sigma_a\\\sigma_b\end{smallmatrix}\right)\right\}$\\
     \hline
     \textbf{boundary} & \textit{hierar. connectivity}: parent boundary-ID\\
     \textbf{(edge)} & \textit{statistics}: boundary length $l$; average contrast $\overline{\delta}$\\
     \hline
    \end{tabular}
   \end{center}
   \end{scriptsize}
   \caption{Data stored per cluster and boundary element.}
   \label{tab:statistics}
  \end{table}
  
  \begin{figure}[t]
   \begin{minipage}[t]{0.38\textwidth}
    \centering
    \includegraphics[height=2.5cm]{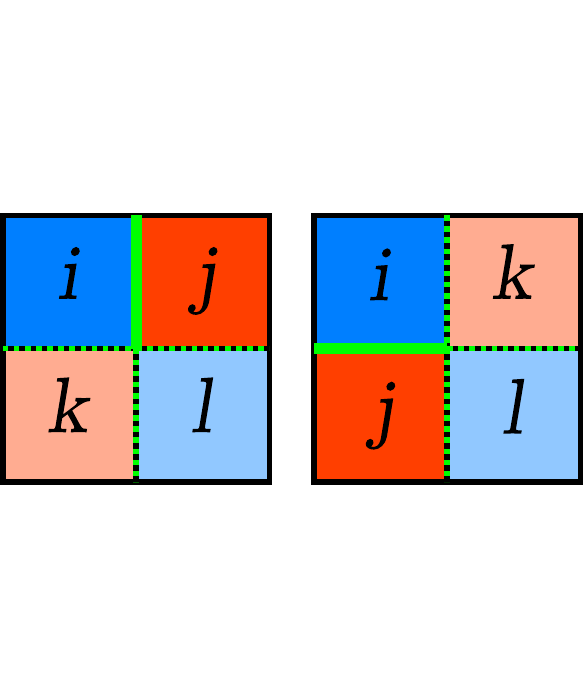}
    \caption{The boundary segment between $i$ and $j$ is a diagonal one, if the labels
    of $i$ with $l$ or $j$ with $k$ are equal.}
    \label{fig:diag_bound_check}
   \end{minipage}\hfill
   \begin{minipage}[t]{0.6\textwidth}
    \centering
    \includegraphics[height=2.5cm]{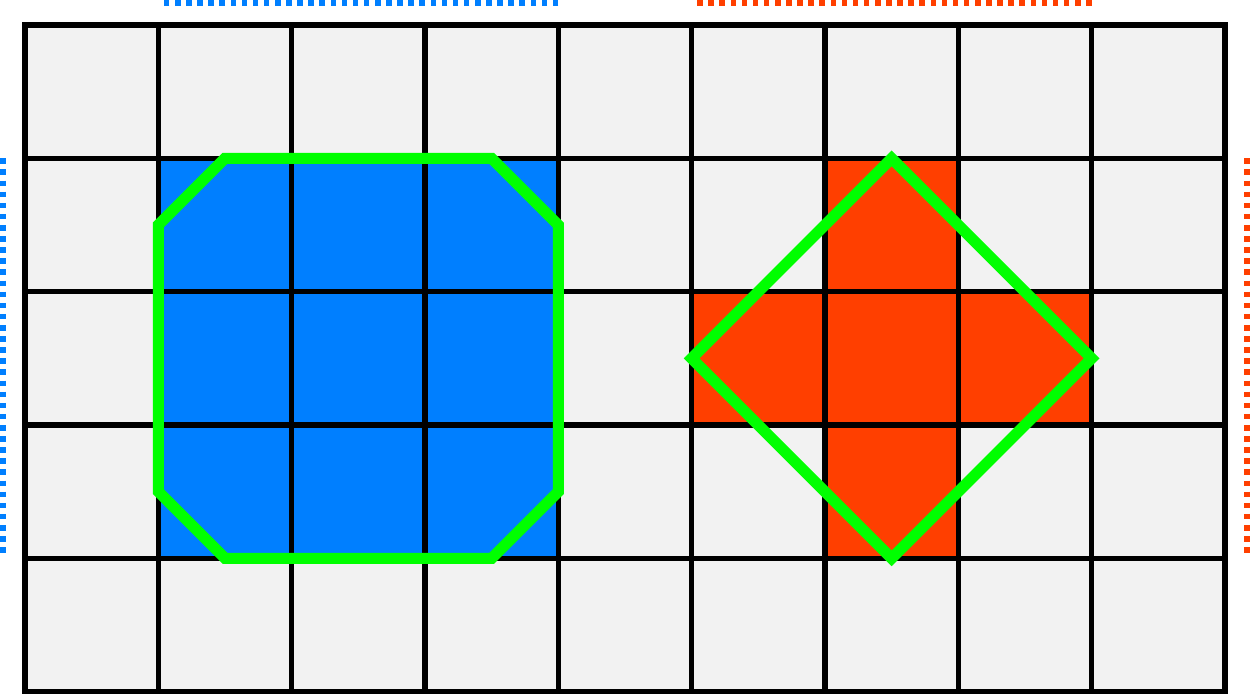}
    \caption{Both blue and red clusters have a boundary length of 12 counting in
    $\mathbf{L}^1$-norm. With shortening, $\mathbf{L}^2$-norm is approximated to length
    10.82 for the blue and 8.49 for the red cluster.}
    \label{fig:L2_norm_short}
   \end{minipage}
  \end{figure}
  The watershed transform uniquely labels each pixels such that it results in a set
  of connected components each denoting an atomic building block of the image.
  In order to setup the graph based hierarchical clustering, we need to gather region
  adjacencies and feature statistics. Our graph structure is made of indexed sets of
  cluster (node) and boundary (edge) objects, each containing the information listed in
  table~\ref{tab:statistics}.
  With a single scan through the image and labels, one could assign this information.
  The hierarchical connectivity is initialized by a constant \textsc{not\_connected}.
  Each pixel joins its corresponding cluster's size and color statistics. If the label
  to the right or bottom  is different from the current, we have to initiate/update
  the corresponding boundary element:
  \begin{itemize}
   \item Let $i$ and $j$ be the labels and w.l.o.g. $i<j$, we look for the first
   neighbor-ID $\geq i$ in the (sorted) adjacency list of cluster $j$. If we found
   $i$, we increase the border length of $b_{ij}$ by 1, else we initiate a new
   boundary of length 1 and add the link ahead of the found position in $j$'s
   adjacency list.
   \item We check the labels below/right for presence of a diagonal boundary (cf.
   Fig.~\ref{fig:diag_bound_check}). If found, we shorten the border length of $b_{ij}$
   by $\nicefrac{\sqrt{2}}{2}$ in order to approximate $\mathbf{L}^2$-norm (cf.
   Fig.~\ref{fig:L2_norm_short}).   
  \end{itemize}
  Then, we iterate the clusters once: for each neighbor in the current cluster's
  adjacency list, we fill in the value of our cluster-distance and insert the backlink
  at the end of the neighbor's adjacency list. Note that this operation retains the
  correct sorting of adjacency lists, because all neighbor-IDs were bigger and the
  clusters are processed by increasing index. Furthermore, we bookmark the nearest
  neighbor per cluster with respect to cluster-distances.
  
  The runtime complexity of this tree foundation is $\mathcal{O}(n)$, since we visit
  each pixel only once and the individual operations are of constant complexity.
  Please note that the degree of a node with respect to neighborhood edges (its
  adjacency list size) is constant on average ($<6$) as a conclusion from
  "Euler's planar graph formula",
  since only segments touching in the image domain become connected by an edge in the graph.

\subsection{Agglomerative Clustering}
\label{ssec:agg_clustering}
  We impose a graph structure of nodes and edges (cf. Tab.~\ref{tab:statistics}) on the
  image. A node represents a cluster (set of connected pixels). There are two kinds
  of edges between clusters:
  the first are planar edges, which represent a common boundary between some pair of
  clusters with respect to 4-connectedness of comprised pixels and when existing at the
  same \textit{time} in the hierarchical tree. In this sense, a node's lifetime spans
  from its creation until it is merged. Thus, the timeline is made of the chronological
  order of merge events in the hierarchy tree.  
  The others are hierarchical edges that connect a pair of merged segments with their
  parent. These hierarchy edges form a binary tree with the root node representing the
  whole scene down to the leaves being atomic clusters resulting from the preceding
  watershed transform.
  
  At each step, agglomerative clustering chooses the cluster pair of lowest distance
  to be merged. Therefore, all current candidate pairs are maintained in a priority
  queue ordered by their distance. For this, we employ a Fibonacci heap implementation.
  Here, the \textsc{delete} operation is most costly with $\mathcal{O}(\log{m})$ runtime
  complexity in the number of heap entries. Since every merged pair has to be deleted
  from the heap once, the overall complexity is dominated by $\mathcal{O}(n\log{m})$.
  Thus, to optimize the runtime, we have to keep the number $m$ of candidates in the
  heap as low as possible. From section \ref{ssec:init_tree} we already know that $m<3n$
  even if we consider all planarly connected clusters. Pushing only the nearest-neighbor
  connected to each cluster to the heap would guarantee $m\leq n$. We can improve this
  further to $m\ll \nicefrac{n}{2}$ pushing only reciprocal nearest neighbor (RNN) pairs
  to the heap, since this is a necessary condition for the pair of
  \textit{globally lowest} distance.
  Unfortunately, a further improvement which caches
  nearest-neighbor-chains\footnote{\url{https://en.wikipedia.org/wiki/Nearest-neighbor_chain_algorithm}}
  (NNC) is not applicable, since the local connectivity violates the
  \textit{reducibility} prerequisite~\cite{bruynooghe77}.
  
  Each time two clusters are merged, a new parent cluster is initialized with the joined
  statistics and adjacencies. Merging of two sorted adjacency lists into one is of linear
  time in the number of
  entries,
  thus amortized constant in this case. Furthermore, this way the case where a cluster
  was neighboring both children can be easily identified and treated in special way: the
  two boundaries between the cluster and both children are concatenated, in other words
  a new boundary object with added lengths and weighted harmonic mean of contrasts is
  created to replace them. The cluster-distances between the parent and its neighbors are
  calculated, nearest-neighbor edges identified and finally corresponding RNNs updated
  in the heap.
  When only a single active cluster is left over, it represents the whole scene and is
  the root of the finalized hierarchy tree.

\subsection{Cluster-Distance from Boundary Contrast and Surface Dissimilarity}
  While handling of data structures is important for efficiency, the quality of the
  hierarchical segmentation is determined by the design of a proper cluster-distance.
  Hence, this is one of our main contributions. Our distance function $\mathcal{D}$ is
  threefold, comprising the following parts:
  \begin{eqnarray}
   \mathcal{D}(P,Q) &=& \log\left(\omega\left(P,Q\right)\right) \quad\;\textit{(surface dissimilarity)}  \\\nonumber
                    &+& \log\left(\overline{\delta}\left(P,Q\right)\right) \quad\;\textit{(boundary contrast)} \\\nonumber
                    &+& \log\left(\eta\left(P,Q\right)\right) \quad\;\textit{(spatial linkage).}
   \label{eqn:clusterdistance}
  \end{eqnarray}
  Here, $P$ and $Q$ denote clusters with a common boundary. All three parts are
  innovative in some aspect. In the following subsections, we will explain each term in
  more detail. Please note that adding the logarithms is a geometric fusion of terms
  (equivalent to a product), hence one does not need to normalize the scales of individual
  parts to a common range. Nonetheless, it is suggestive to train weights for an
  optimized combination of parts. This could further improve results at virtually no
  costs, but was not yet exploited for experiments in this paper.
  
\noindent\textbf{Surface Dissimilarity Term}
  Bucking the trend, our approach is very thrifty in computing different kinds of
  features. Indeed, we only use a normal distribution of colors to describe the
  appearance of a cluster (cf. Tab.~\ref{tab:statistics}).
  These statistics are gathered in the form of ML-estimates from the individual colors
  of all pixels belonging to a certain cluster. We primarily came to this decision,
  because the joining of Gaussian statistics is a very efficient, constant time
  operation~\cite{Chan83}.  
  The second good reason is that the Wasserstein distance between normal distributions
  is meaningful and fast to compute. The Wasserstein distance is a transport metric
  between probability distributions.
  It accounts how much (probability-) mass needs to be carried how far wrt. an
  underlying metric space. Here, the underlying space is CIE-Lab with
  $\mathbf{L}^1$-norm\footnote{This is similar to the well known earth mover's distance
  (EMD) on histograms, which is in fact the discretized $\mathcal{W}_1$ distance.},
  which mimics human texture discrimination in a simplified way when working on top
  of the perceptually normalized CIE-Lab colorspace~\cite{rubner2000emd}.
  Using normal distributions, the 1st Wasserstein distance solves to the expression 
  \begin{eqnarray}
   \radigwdist(\mathcal{N}\!_P,\mathcal{N}\!_Q) &=& \radigabs{\vec{\mu}_P - \vec{\mu}_Q}
    + \radigabs{\radigtrace{\sqrt{\Sigma_P}} - \radigtrace{\sqrt{\Sigma_Q}}} .
  \label{eqn:w1distGauss}
  \end{eqnarray}

  Ward's clustering criterion~\cite{ward1963} is a distance which estimates
  the growth in data variance when joining clusters,
  \begin{equation}
   \operatorname{Ward}(P,Q) = \frac{ d\!\left(\mu_P, \mu_Q\right)^2}{\nicefrac{1}{A_P} + \nicefrac{1}{A_Q}}
   \label{eqn:ward}
  \end{equation}
  where $A$ refers to the clusters' surface areas in pixels.
  If you think of this in terms of color means, the intuition is that mixing more
  and more colors always ends up in some grayish tone, but then subtle differences
  between such mashes can still make a clear difference for large surfaces.
  This measure is known for developing clusters of more balanced sizes. As a
  novelty, we propose to lift this concept from color to texture by replacing the
  distance between means with the 1st Wasserstein distance, which yields
  \begin{equation}
    \omega(P,Q) = \frac{\radigwdist(\mathcal{N}\!_P,\mathcal{N}\!_Q)^2}{\nicefrac{1}{A_P} + \nicefrac{1}{A_Q}}.
   \label{eqn:ward_wasserstein}
  \end{equation}
  
\noindent\textbf{Boundary Contrast Term}
  Each boundary element carries information about its average (weighted harmonic mean) boundary contrast
  $\overline{\delta}(P,Q)$ between the clusters on either side. This value is
  kept up-to-date if two boundary elements are concatenated during a merge event (cf.
  Sec.~\ref{ssec:agg_clustering}). The remaining question is how it should be
  initialized. Instead of using the gradient magnitude or related measures on the
  contour at pixel resolution, we decided to plug in the Wasserstein distance between
  $P$ and $Q$. Obviously, this is most economic since it is already computed as a part of
  the surface dissimilarity. But there are more good reasons for this decision: a measure
  incorporating the area alongside the contour-line better assesses the strength of
  blurry edges often emerging from motion or defocussing. Furthermore, we smartly avoid
  cross-talk artifacts at the crossings of edges, a problem Arbel\'aez
  et.~al.~\cite{arbelaez2011contour,arbelaez2014mcg} explicitly addressed by a procedure
  called Oriented Watershed Transform (OWT). Please note that despite re-using the
  Wasserstein distance, the boundary contrast is essentially different to the surface
  dissimilarity: from the way boundaries and clusters are merged, the former is the
  average difference between the opposing, atomic clusters along a boundary, while the
  latter includes the difference between both entire surfaces.

\noindent\textbf{Spatial Linkage Term}
  From graph topology, it is a binary decision if two segments are connected by an
  edge or not. However, it is reasonable to consider a certain factor of how close
  they actually are. We introduce the simple yet effective connectivity term
  \begin{equation}
   \eta(P,Q)
   = \left( \frac{\sqrt{A_P}}{l_{PQ}} \cdot
    \frac{\sqrt{A_Q}}{l_{PQ}}\right)^{\nicefrac{1}{2}}
   = \frac{(A_P \cdot A_Q)^{\nicefrac{1}{4}}}{l_{PQ}},
  \label{eqn:spatial_connectivity}
  \end{equation}
  which combines the extents of the common boundary and both surfaces.
  This way, the linkage between clusters is rated independently of concrete positions
  and thus is very flexible in shape. Relating the radical of the cluster surface to the
  common boundary's length is a normalized measure of connectivity, but advances versus
  simply using the perimeter by preferring more convex, smooth shapes versus elongated
  or jagged ones. We put the geometric mean of both surfaces' score, since it
  devaluates a cluster pair more towards the inferior result and thus better avoids
  residues staying alive.

\section{Evaluation of Key Contributions}
  \label{sec:eval_contrib}
    \begin{figure}[t]
    \begin{center}
      \includegraphics[width=0.75\linewidth]{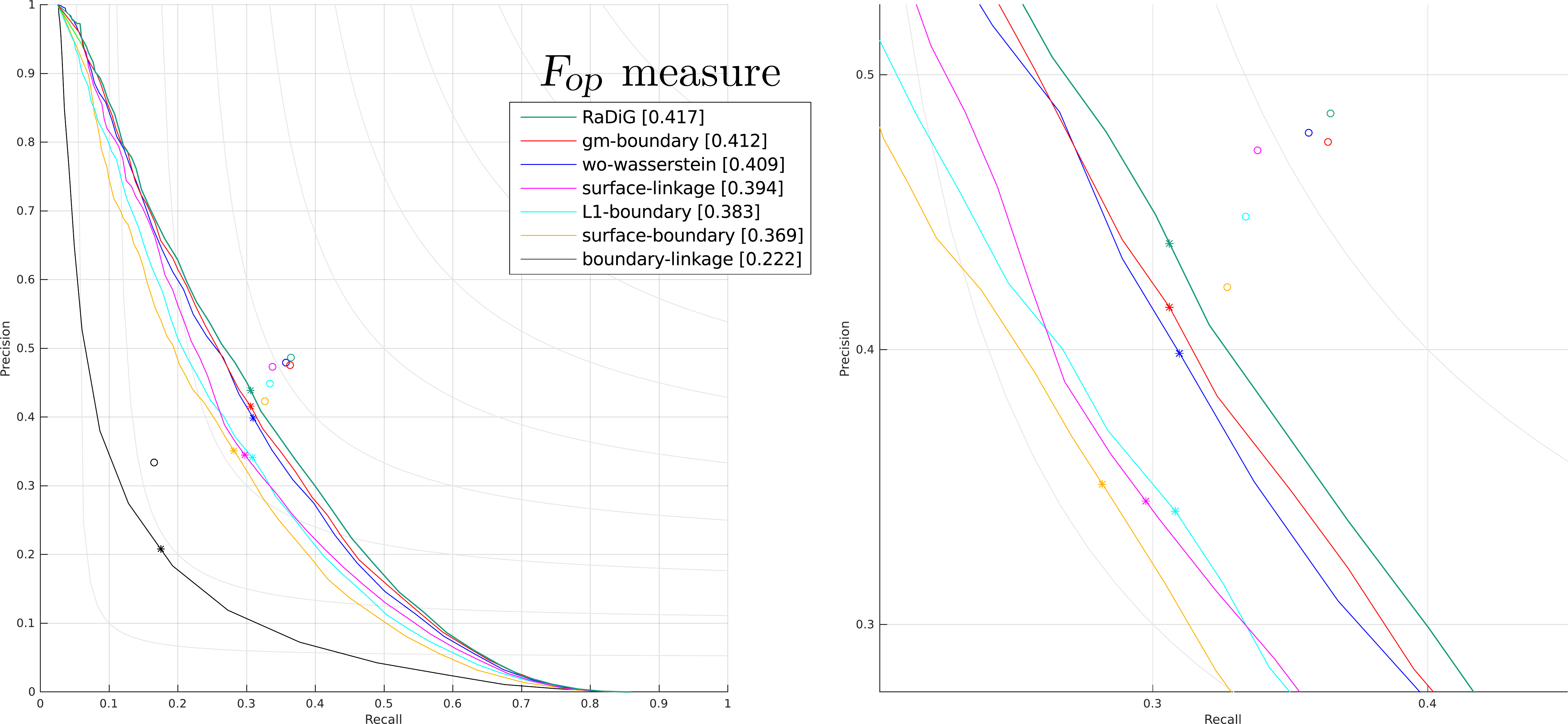}\vspace{-4mm}
    \end{center}
       \caption{Precision/recall curve of RaDiG and its reduced variants on BSDS500
       trainval set using the object-and-parts $F_{op}$ measure. The optimal dataset
       scale (ODS) is drawn with an asterisk on each curve, while the optimal per image
       scale (OIS) is drawn with a circle and numbers are given in the legend.}
    \label{fig:eval_contrib}
    \end{figure}
  Since we introduced several contributions in this paper, it is reasonable to
  evaluate how much the overall result depends on every single innovation. Therefore,
  we compare different variants of our system each with one enhancement deactivated
  or replaced. The experiments are conducted on the trainval images of the BSDS500
  benchmark~\cite{arbelaez2011contour}. This dataset provides boundary annotations of
  several human subjects, which where interpreted to define a segmentation.

  \noindent\textbf{Segmentation Quality Measure}
  It is not straightforward to rate the
  quality of a segmentation with respect to a ground truth. Pont-Tuset and Marques~\cite{pont2013measures}
  defined the $F_{op}$ measure in the context of generating object or part candidates
  for recognition. Basically, this allows a segment to be fragmented into several
  parts. Regions need to exceed some relative overlaps to be accepted as valid
  object resp. part matches. Those matches are weighted by their fraction of overlap,
  thus, parts are only partially counted as true positive matches. On this basis, a
  precision and recall for object-and-part matches is defined. Varying a threshold
  on the cluster-distance of the cluster hierarchy (or UCM) produces different levels
  of segmentations from coarse to fine. Coarse segmentations are likely of low
  precision, but high recall, and vice versa comparing fine oversegmentations. The
  $F_{op}$ measure is the maximally achievable value of the harmonic mean of these
  precision and recall pairs.
  
  \noindent\textbf{Boundary Contrast Evaluation} The RaDiG boundary contrast is initialized
  with the Wasserstein distance between neighboring, atomic clusters and further on
  averaged with respect to approximated $\mathbf{L}^2$-norm length when two parts are
  concatenated. To assess if our novelties really contribute to the overall results,
  we conducted two experiments:  
  Figure~\ref{fig:eval_contrib} shows the original algorithm's result (RaDiG) in
  comparison to a variant where we averaged the gradient magnitude along the
  boundary in order to initialize the boundary contrasts (gm-boundary), which is
  more expensive to compute but slightly inferior.
  We also tried $\mathbf{L}^1$-norm lengths (Fig.~\ref{fig:eval_contrib}, L1-boundary)
  for averaging the boundary contrast, but here the little additional effort for approximation pays off
  well.
  
  \noindent\textbf{Appearance Distance Evaluation} Next, we assess the benefit of integrating
  covariance information using the Wasserstein distance. Thus, we replace it with
  usual squared Euclidean metric between the means, yielding the classic Ward
  clustering criterion and a simpler boundary contrast (Fig.~\ref{fig:eval_contrib},
  wo-wasserstein).
  We deduce that the Wasserstein distance is a worthwhile enhancement, raising the $F_{op}$ from $0.409$ to $0.417$ and showing a consistently better trend on the precision/recall curves. From our observation, the difference is pronounced with larger segments, when textures are more often of non-uniform color.
  
  \noindent\textbf{Threefold Cluster-Distance Evaluation} Our agglomerative clustering approach
  efficiently maintains statistics from both the boundaries between and the surfaces
  of clusters, which gives rise to our sophisticated, threefold cluster-distance term
  (cf. Eq.~(\ref{eqn:clusterdistance})).
  In this experiment, we figure out if all three components contribute to the fine
  overall results, by omitting one of the parts at a time. We named the twofold
  cluster-distances by the remaining components: surface-linkage, surface-boundary, and
  boundary-linkage (cf. Fig.~\ref{fig:eval_contrib}). Dropping one of the parts severely
  worsens the results. Particularly important is the Ward-based term of surface
  dissimilarity, probably because it controls the clusters' sizes not to diverge too much
  during agglomeration.
  
  \begin{table}[t]
   \begin{minipage}[b]{0.48\textwidth}
    \centering
    \begin{scriptsize}
    \begin{tabular}{|c|c|c|c|}
      \hline
      \textbf{resolution} & \textbf{\#pixel} & \textbf{time in ms} \\
      \hline
      $135\times90$ & $12150$ & $11$ \\
      $270\times180$ & $48600$ & $41$ \\
      $320\times240$ & $76800$ & $64$ \\
      $481\times321$ & $154401$ & $107$ \\
      $540\times360$ & $194400$ & $151$ \\
      $1080\times720$ & $777600$ & $604$ \\
      \hline
    \end{tabular}
    \end{scriptsize}
    \caption{Runtime of RaDiG on a single core of an Intel Xeon X5680 @ 3.33GHz for 
    different resolutions.}
    \label{tab:runtime}
   \end{minipage}\hfill
   \begin{minipage}[b]{0.48\textwidth}
    \centering
    \begin{scriptsize}
    \begin{tabular}{|c|c|c|c|}
      \hline
      \textbf{component} & \textbf{time in ms} & \textbf{time in \%} \\
      \hline
      colorspace & $10$ & $16$ \\
      watershed & $7$ & $10$ \\
      founding tree & $20$ & $32$ \\
      agg. clustering & $27$ & $42$ \\
      \hline
      \textbf{$\sum$serial} & $64$ & $100$ \\
      \textbf{$\sum$parallel} & $27$ & $42$ \\
      \hline
    \end{tabular}
    \end{scriptsize}
    \caption{Runtime of different components of our RaDiG approach at QVGA resolution ($320\times240$).}
    \label{tab:runtime2}
   \end{minipage}
  \end{table}
  
  \noindent\textbf{Runtime Evaluation} We claimed that our implementation is realtime
  capable and therefore the best choice for applications on mobile platforms such as
  autonomous robots.
  Table~\ref{tab:runtime} shows the overall runtime of our approach for different
  resolutions, estimated as an average from six cluttered, natural images. Empirically,
  it seems as if, thanks to pruning of candidate pairs in the heap data structure (cf.
  Sec.~\ref{ssec:agg_clustering}), we achieved an amortized linear behavior in the
  number of pixels. For QVGA resolution, table~\ref{tab:runtime2} breaks down how much
  time each component of our approach consumes. Since our implementation is based on
  the ROS-framework~\cite{ros} and every component is implemented as a processing node
  using its own thread, only the slowest component restricts the framerate when
  processing video data. Thus we are able handle up to 37\,fps in QVGA on our machine. 
  On BSDS500 image size ($481\times321$), the single thread variant of RaDiG takes on
  average $107$ms to run the whole algorithm. Thereby, it outperforms the approaches of
  Arbel\'aez et.~al.~\cite{arbelaez2014mcg} by one resp. two orders of magnitudes. Testing their code of
  version 2.0 on the same machine, we measured runtimes of $17.6$ seconds for MCG-UCM ($165\times$ slower) and $2.8$ seconds for SCG-UCM ($26\times$ slower) on BSDS500 data.

\section{Comparative Experiments on BSDS500 and Pascal-Context}
\label{sec:eval_comp}
\begin{figure}[t]
  \begin{center}
    \includegraphics[width=0.155\linewidth]{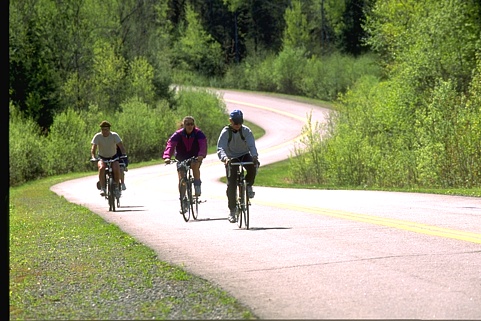}
    \includegraphics[width=0.155\linewidth]{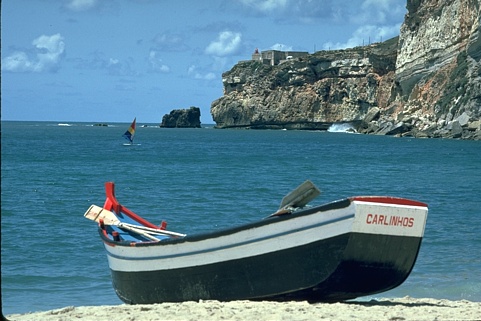}
    \includegraphics[width=0.155\linewidth]{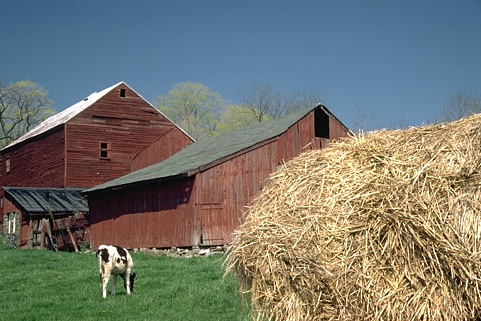}
    \includegraphics[width=0.155\linewidth]{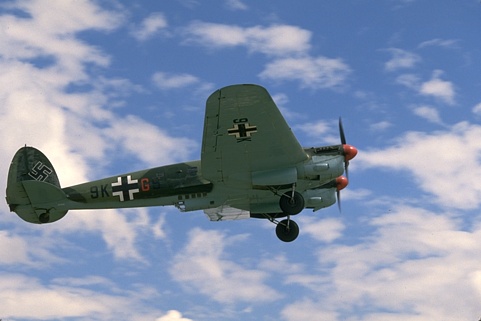}
    \includegraphics[width=0.11\linewidth]{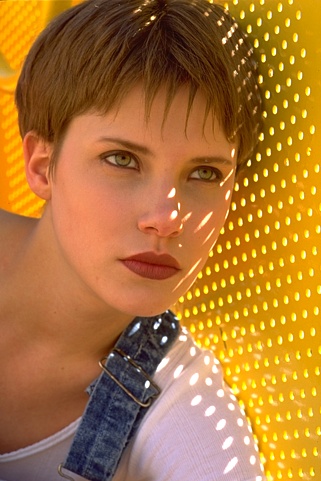}
    \includegraphics[width=0.11\linewidth]{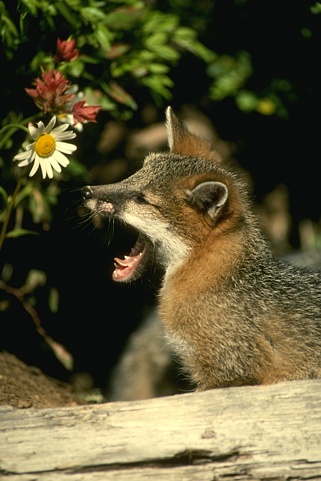}
    \includegraphics[width=0.11\linewidth]{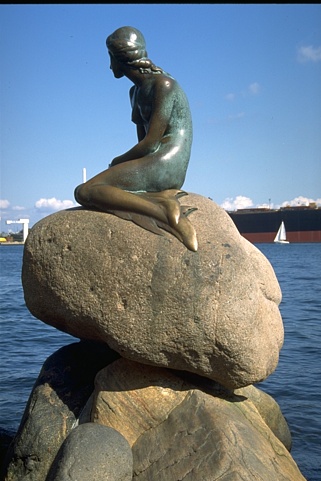}\\ 
    \includegraphics[width=0.155\linewidth]{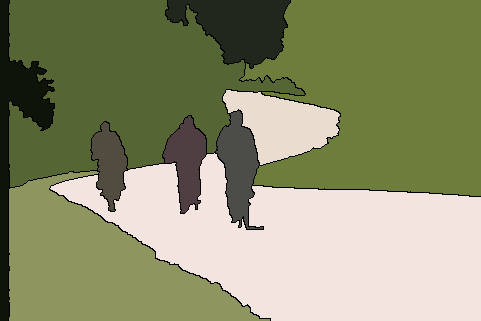}
    \includegraphics[width=0.155\linewidth]{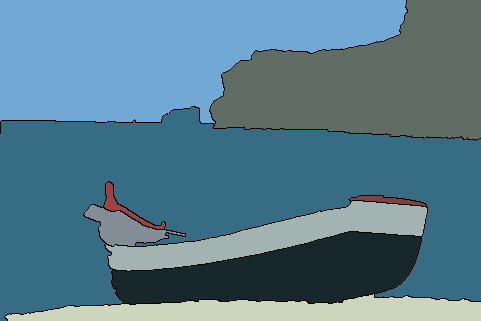}
    \includegraphics[width=0.155\linewidth]{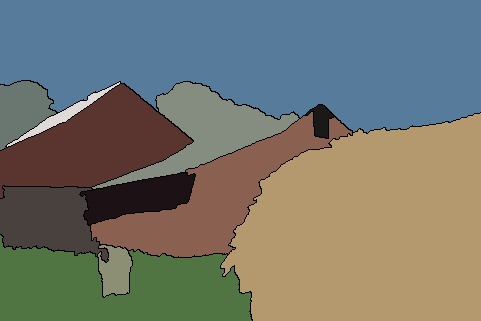}
    \includegraphics[width=0.155\linewidth]{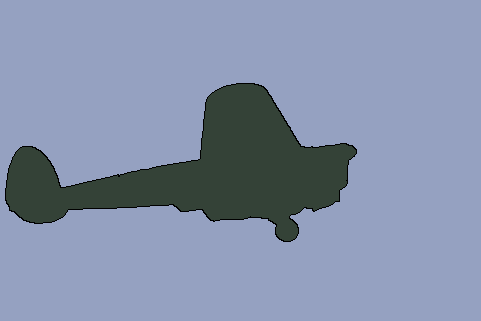}
    \includegraphics[width=0.11\linewidth]{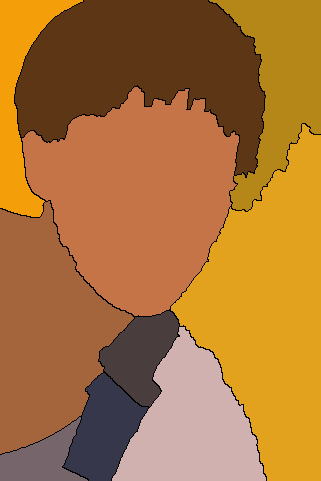}
    \includegraphics[width=0.11\linewidth]{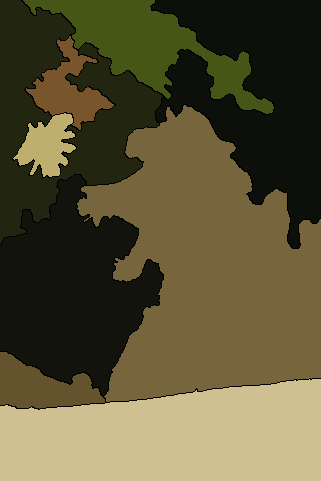}
    \includegraphics[width=0.11\linewidth]{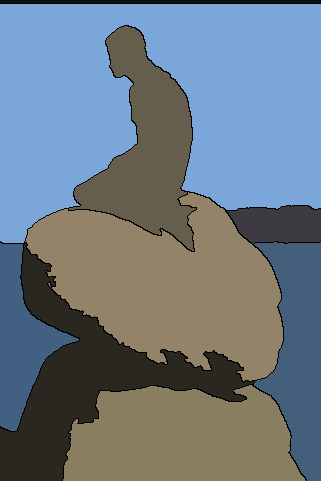}\\ 
    \includegraphics[width=0.155\linewidth]{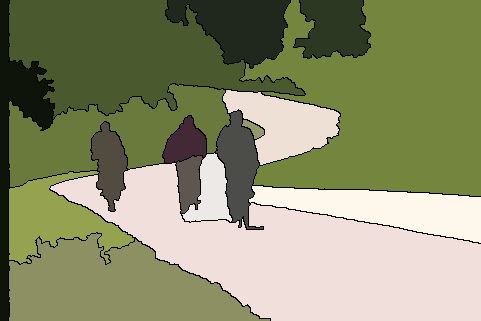}
    \includegraphics[width=0.155\linewidth]{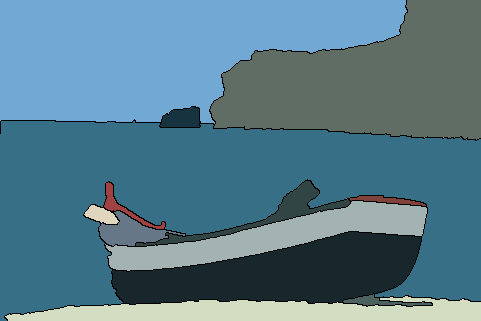}
    \includegraphics[width=0.155\linewidth]{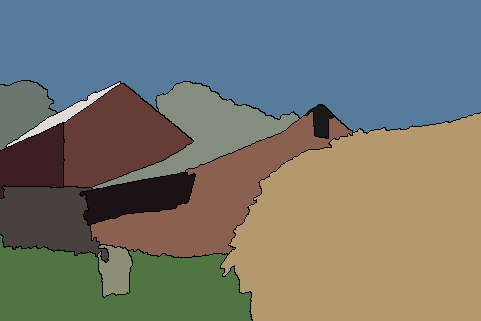}
    \includegraphics[width=0.155\linewidth]{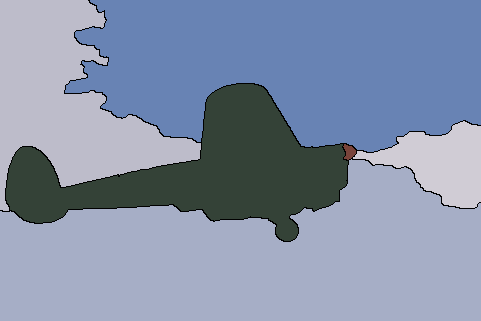}
    \includegraphics[width=0.11\linewidth]{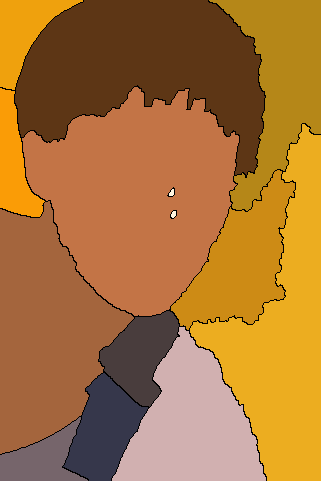}
    \includegraphics[width=0.11\linewidth]{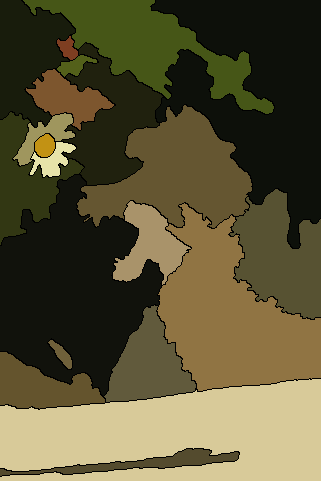}
    \includegraphics[width=0.11\linewidth]{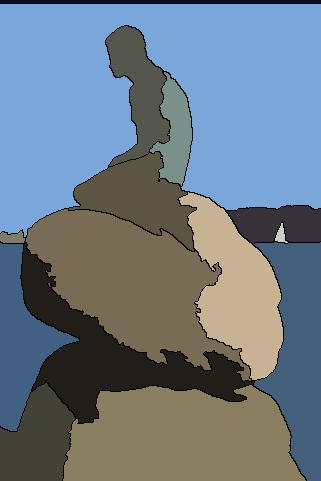}\\ 
    \includegraphics[width=0.155\linewidth]{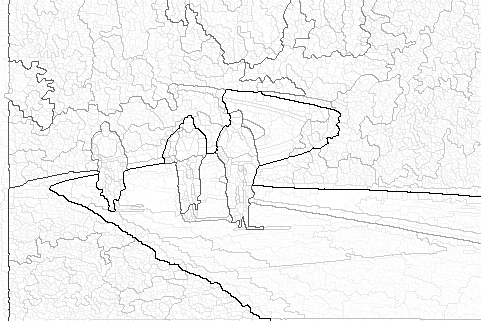}
    \includegraphics[width=0.155\linewidth]{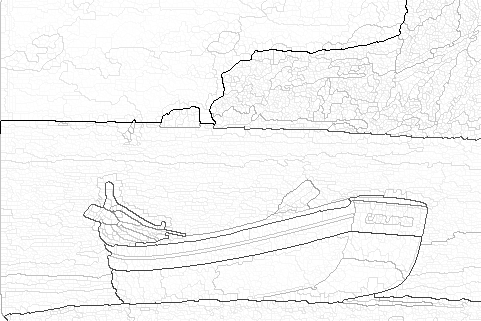}
    \includegraphics[width=0.155\linewidth]{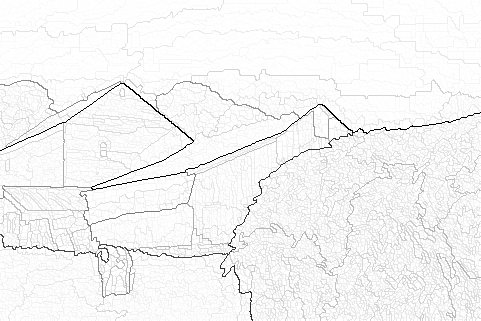}
    \includegraphics[width=0.155\linewidth]{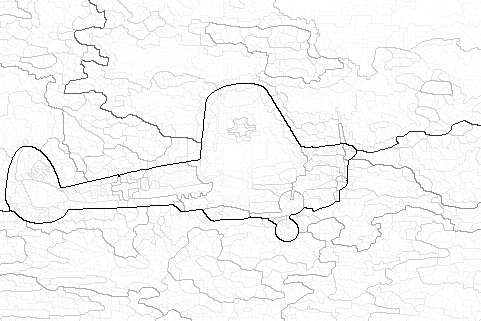}
    \includegraphics[width=0.11\linewidth]{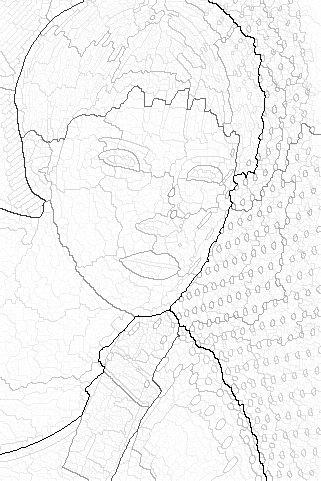}
    \includegraphics[width=0.11\linewidth]{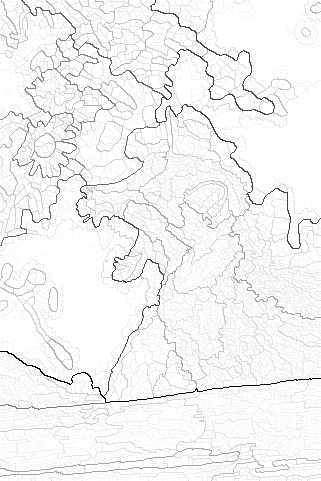}
    \includegraphics[width=0.11\linewidth]{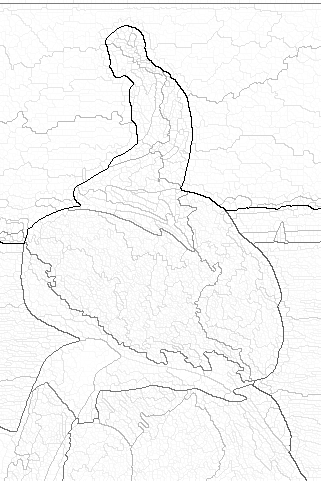}\vspace{-4mm}
  \end{center}
     \caption{Exemplary segmentations of RaDiG on BSDS500 images. From top to bottom per column: original, optimal image scale (OIS), optimal dataset scale (ODS), and ultrametric contour map (UCM) of the corresponding hierarchy.}
  \label{fig:bsds_examples}
  \end{figure}

  We compared absolute performances and ranked our approach versus state-of-the-art ones
  on widely used datasets and appropriate measures.
  Pont-Tuset and Marques~\cite{pont2013measures} provide a number of results from
  different algorithms based on their $F_{op}$ measure. In addition, we wrote a tool
  to convert data from the result formats of Arbel\'aez et.\ al.\ \cite{arbelaez2014mcg}
  plus Ren and Shakhnarovich~\cite{ren2013iscra} to update their comparison with
  recent approaches. We are convinced that $F_{op}$ is the most adequate performance
  metric for our kind of application, since we aim to perceive semantically meaningful
  entities. Our method is not an edge detector in the first place, nevertheless we
  include the $F_b$ measure, which quantifies how well contours are reproduced, as
  additional assessment.
    \begin{figure}[t]
    \begin{center}
      \includegraphics[width=0.75\linewidth]{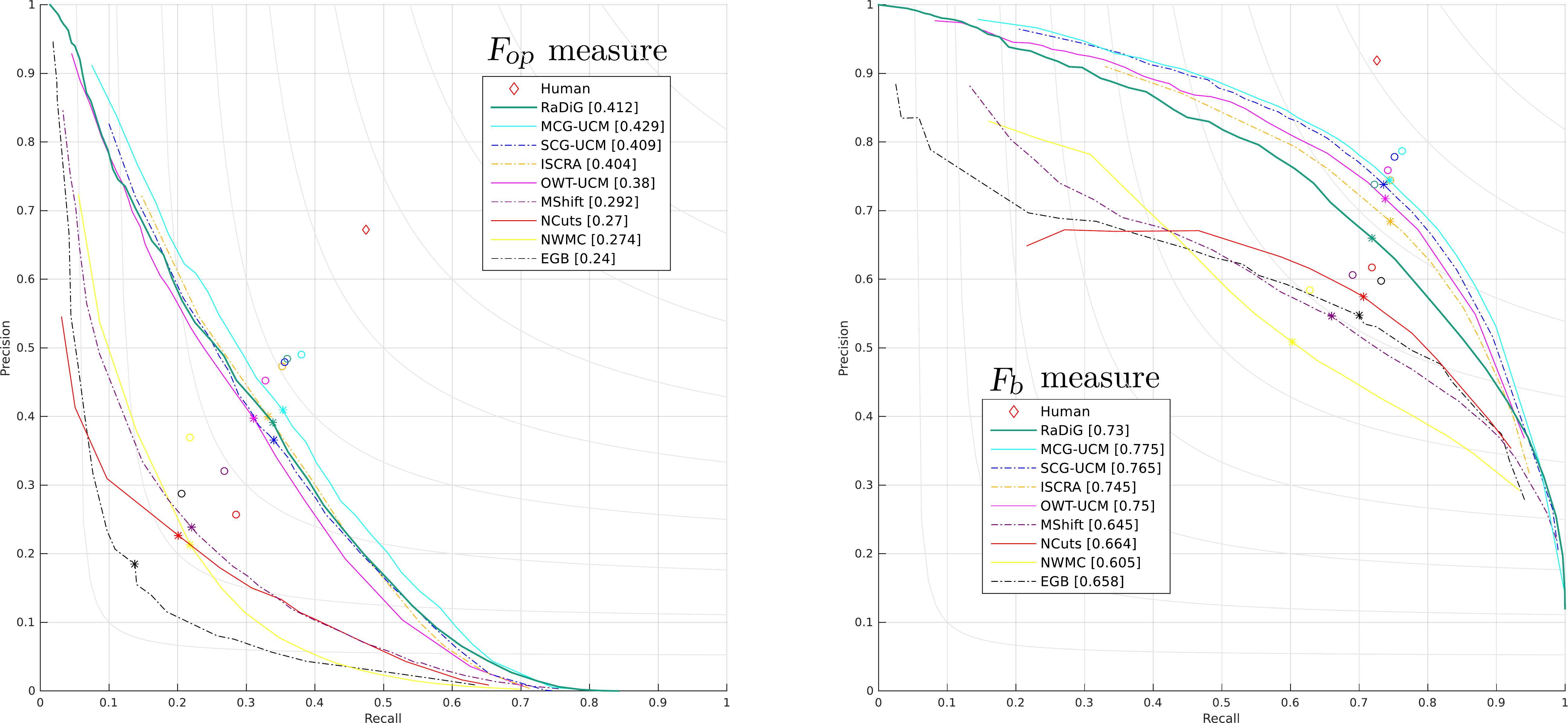}\vspace*{2mm}\\
      \includegraphics[width=0.75\linewidth]{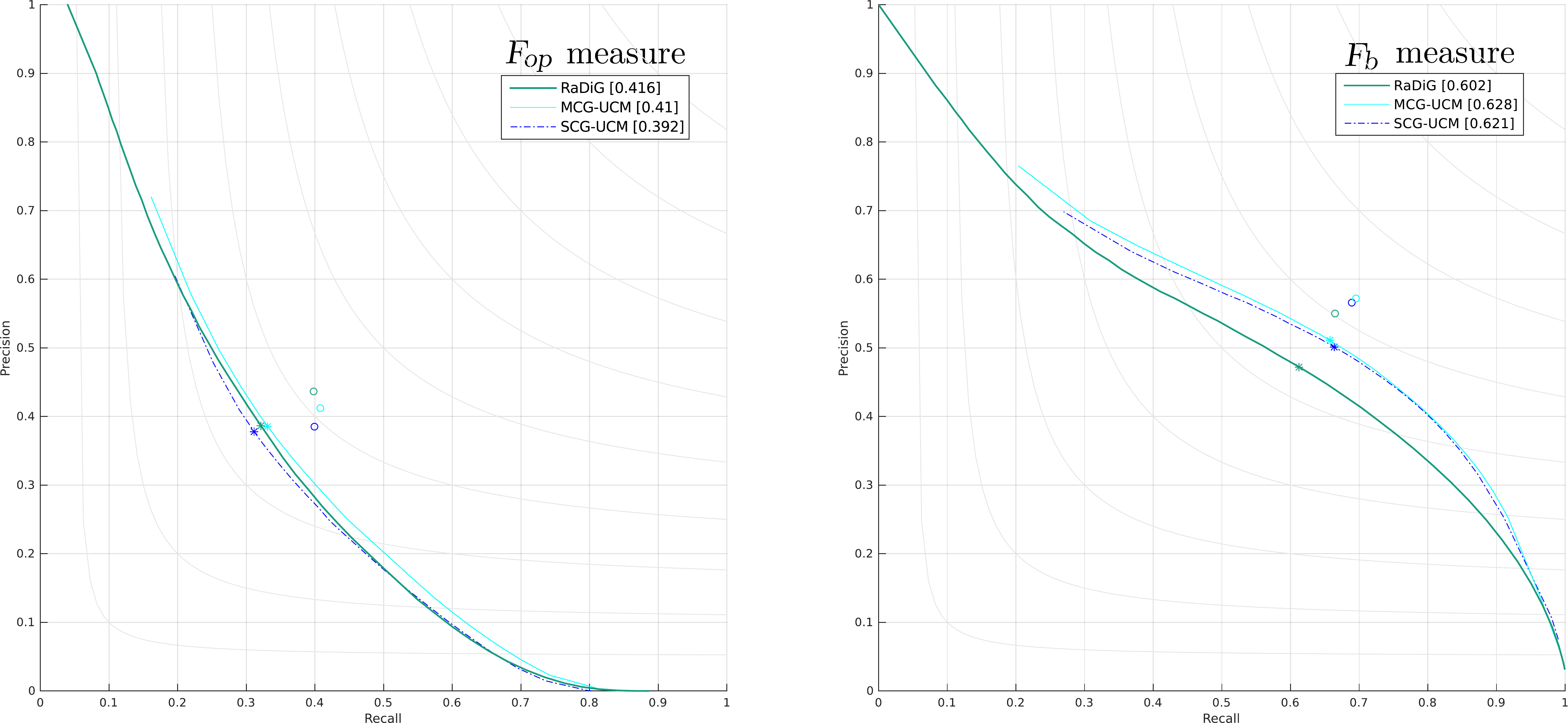}
    \end{center}
       \caption{Precision/recall curves of RaDiG and leading approaches using $F_{op}$ as well as $F_{b}$ measure. Top row: results on BSDS500 test images; second row: results on Pascal-Context dataset. The optimal per image scale (OIS) values (circles) are given in the legend.}
    \label{fig:bsds500_pascalc_comp}
    \end{figure}
  Figure~\ref{fig:bsds500_pascalc_comp} first row shows the precision/recall curves for $F_{op}$ and $F_b$ measures of all tested approaches on the test images of BSDS500. There is a noticeable gap between EGB (efficient graph
  based method of Felzenszwalb and Huttenlocher~\cite{FelzHutt04}), NCUTs (normalized
  cuts~\cite{Cour05}), MShift (mean shift~\cite{Comaniciu2002}), and NWMC (binary partition tree~\cite{vilaplana2008})
  on one hand,
  but the recent approaches OWT-UCM (global probability of boundary with oriented watershed transform followed by agglomerative merging~\cite{arbelaez2011contour}),
  ISCRA (image segmentation by cascaded region agglomeration~\cite{ren2013iscra}),
  SCG-UCM (single-scale combinatorial grouping~\cite{arbelaez2014mcg})
  MCG-UCM (multi-scale combinatorial grouping~\cite{arbelaez2014mcg}) and our proposed
  RaDiG method on the other hand. A clear difference between OWT-UCM, ISCRA, SCG-UCM
  and MCG-UCM versus the others is that they tuned certain aspects of their algorithms
  using machine learning techniques. Also, they are rather costly to compute and do not
  target realtime applications. This said, results of RaDiG are very encouraging
  and our algorithm is presumably the best choice for realtime tasks in unconstrained
  environments, such as often present in autonomous robotics. Figure~\ref{fig:bsds_examples}
  shows some characteristic outcome of RaDiG on BSDS500 images.
  
\begin{figure}[t]
  \begin{center}
    \includegraphics[width=0.16\linewidth]{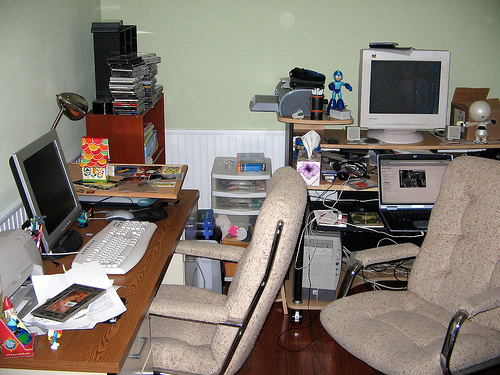}
    \includegraphics[width=0.16\linewidth]{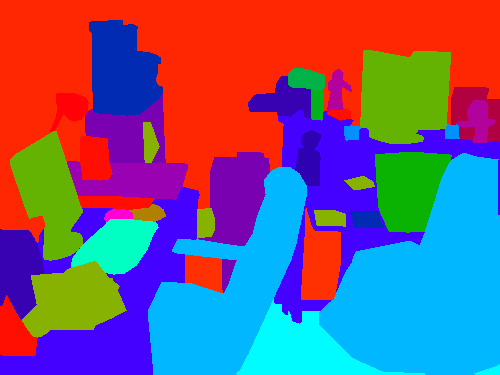}
    \includegraphics[width=0.16\linewidth]{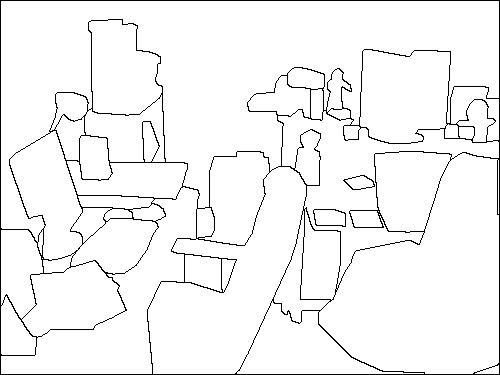}
    \includegraphics[width=0.16\linewidth]{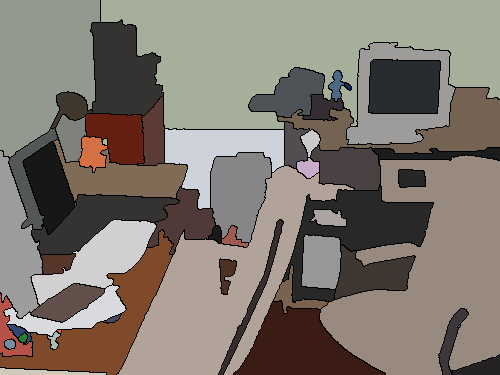}
    \includegraphics[width=0.16\linewidth]{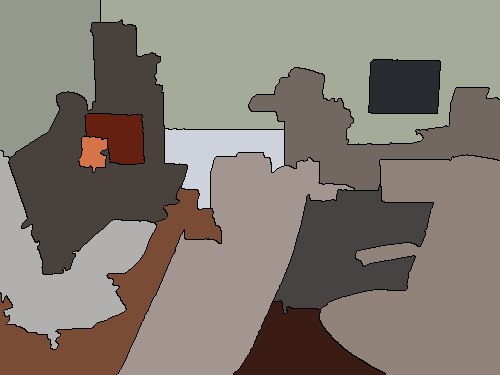}
    \includegraphics[width=0.16\linewidth]{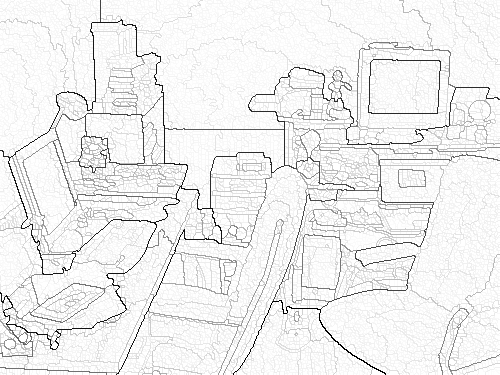}\\
    \includegraphics[width=0.16\linewidth]{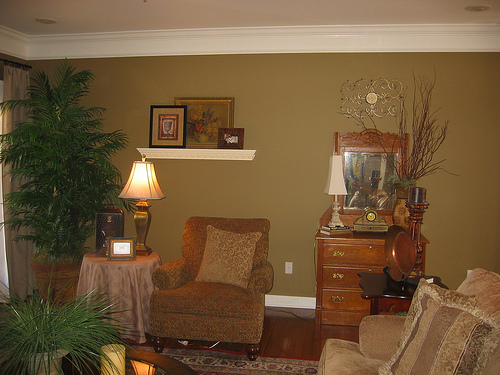}
    \includegraphics[width=0.16\linewidth]{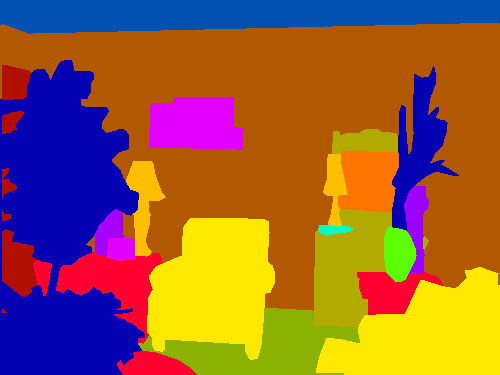}
    \includegraphics[width=0.16\linewidth]{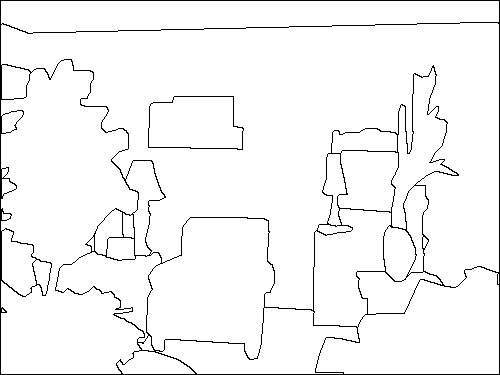} 
    \includegraphics[width=0.16\linewidth]{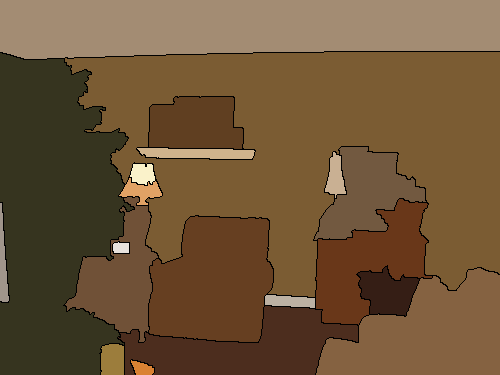}
    \includegraphics[width=0.16\linewidth]{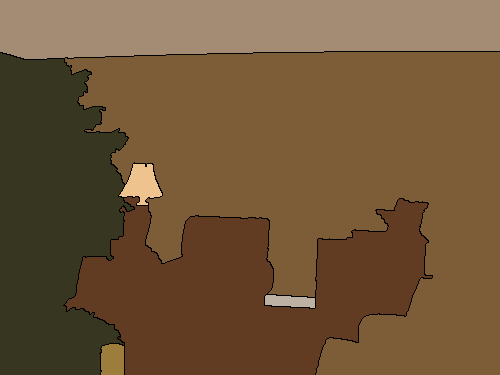}
    \includegraphics[width=0.16\linewidth]{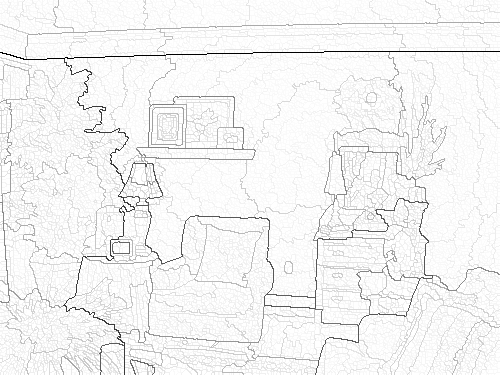}\\
    \includegraphics[width=0.16\linewidth]{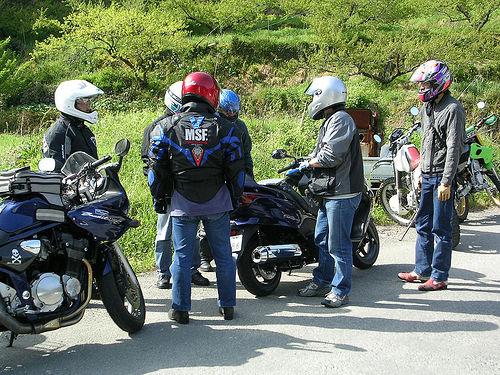}
    \includegraphics[width=0.16\linewidth]{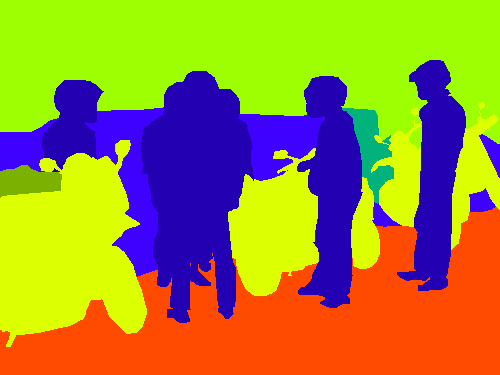}
    \includegraphics[width=0.16\linewidth]{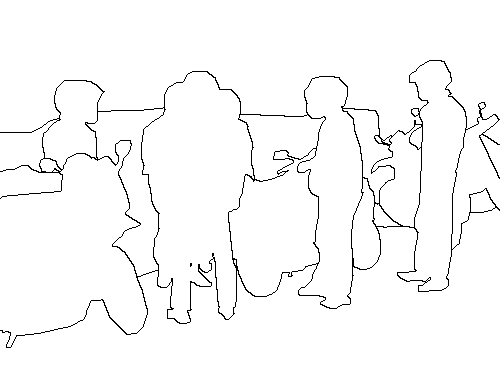} 
    \includegraphics[width=0.16\linewidth]{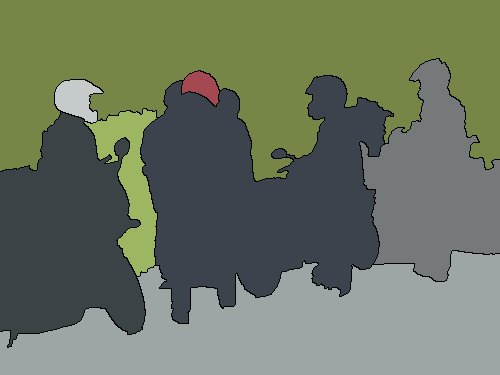}
    \includegraphics[width=0.16\linewidth]{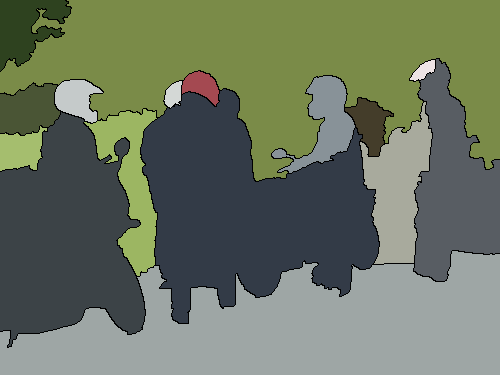}
    \includegraphics[width=0.16\linewidth]{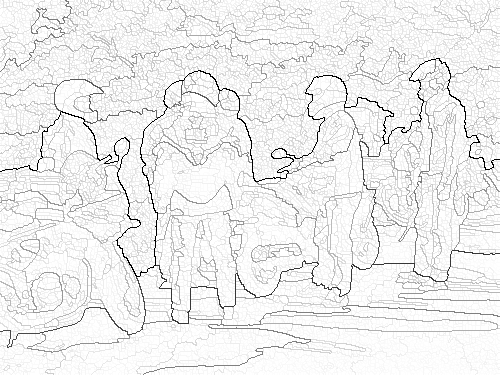}
  \end{center}
     \caption{Exemplary segmentations of RaDiG on Pascal-Context images. From left to right per row: original, ground truth labels, g.t. segmentation, optimal image scale (OIS), optimal dataset scale (ODS), and ultrametric contour map (UCM).}
  \label{fig:pascalc_examples}
  \end{figure}

  While the BSDS500 dataset is the popularly accepted benchmark in this kind of
  unsupervised image segmentation, we felt that the ground truth is not entirely
  appropriate for evaluation of object and part candidates. The Pascal-Context dataset~\cite{mottaghi2014role}
  contains pixel-precise labels for more than 400 object categories on the train/val
  Pascal VOC 2010 data~\cite{pascal-voc-2010} (10103 images), which is appealing also for comparison of
  unsupervised approaches due to this variety and size. We wrote a tool to convert this
  ground truth labels into segmentations by separating labels of the same category
  into spatially connected components, numbering them increasingly and convert them
  into the ground truth format of BSDS500. Fortunately, two of the best performing
  approaches on BSDS500, SCG-UCM and MCG-UCM, published their ultrametric contour maps online\footnote{http://www.eecs.berkeley.edu/Research/Projects/CS/vision/grouping/mcg/} for the
  main Pascal VOC 2012 images, which is a superset of 2010, so that we are able to
  compare results.
  All three approaches retain their very good performance on this large dataset (cf.
  Fig.~\ref{fig:bsds500_pascalc_comp}, second row). However, RaDiG supersedes SCG-UCM on the precision/recall
  curve and the gap between MCG-UCM and RaDiG is even smaller than on BSDS500.
  Furthermore, looking at $F_{op}$ from optimal image scale, RaDiG took the lead.
  However, RaDiG is only moderately suited for precise contour detection, as $F_b$
  results suggest.
  Table~\ref{tab:ODS} reports numbers for the optimal dataset scale
  results of all approaches, which are also marked in the Figures~\ref{fig:eval_contrib} and \ref{fig:bsds500_pascalc_comp} by asterisks.
  Figure~\ref{fig:pascalc_examples} shows some qualitative results
  of RaDiG on Pascal-Context dataset.
  \begin{table}[t]
  \centering
    \begin{scriptsize}
      \begin{tabular}{|l|c|c|c|c|}
        \hline
         & \multicolumn{2}{c|}{\textbf{BSDS500}} & \multicolumn{2}{c|}{\textbf{Pascal-C}}\\
         \,\textbf{method} & $F_{op}$ & $F_{b}$ & $F_{op}$ & $F_{b}$ \\
        \hline
        \,RaDiG & \,$0.363$\, & \,$0.688$\, & \,$0.351$\, & \,$0.533$\,\\
        \,MCG-UCM & $0.379$ & $0.744$ & $0.356$ & $0.575$\\
        \,SCG-UCM & $0.352$ & $0.737$ & $0.342$ & $0.571$\\
        \,ISCRA & $0.363$ & $0.714$ & - & - \\
        \,OWT-UCM\, & $0.349$ & $0.727$ & - & -\\
        \,MShift & $0.229$ & $0.598$ & - & - \\
        \,NCuts & $0.213$ & $0.634$ & - & - \\
        \,NWMC & $0.215$ & $0.552$ & - & - \\
        \,IIDKL & $0.186$ & $0.575$ & - & - \\
		\hline
      \end{tabular}
     \end{scriptsize}
     \caption{Optimal dataset scale(ODS) results for both comparative experiments.}
     \label{tab:ODS}
  \end{table}
  
\section{Conclusion and Future Work}
\label{sec:conclusion}
  We proposed RaDiG (Rapid Distributions Grouping), a fast and greedy algorithm to construct a
  tree of nested image segmentations starting from an atomic subdivision. Whereas its
  overall clustering framework is approved in awhile, we contributed several valuable
  findings. The feature computations and handling of the graph structure were optimized to enable a very low runtime needed for realtime applications.
  At the core of the approach, a novel, threefold cluster-distance was introduced and
  stepwise shown to advance the quality of clustering. Overall, our method plays in the
  same league wrt. precision/recall as current state-of-the-art approaches, but is at
  least an order of magnitude faster.
  By now, RaDiG is a parameter free approach. We are confident that involving machine
  learning techniques to a weighted balancing of components could further improve
  results. At the moment, there are several projects under development, where we deploy
  RaDiG on mobile robots, e.g. in the area of object manipulation.



\bibliographystyle{abbrv}
\bibliography{my}

\end{document}